%% file: main.tex
\documentclass[10pt,twocolumn,letterpaper]{article}

\usepackage[pagenumbers]{cvpr} 

\input{preamble}
\definecolor{cvprblue}{rgb}{0.21,0.49,0.74}
\usepackage[pagebackref,breaklinks,colorlinks,allcolors=cvprblue]{hyperref}
\usepackage{natbib}
\usepackage{xcolor} 
\usepackage{color, colortbl}
\usepackage{arydshln}
\usepackage{fontawesome}

\usepackage[ruled,vlined]{algorithm2e}
\usepackage{amsmath}
\usepackage{times}
\usepackage{amsmath, amsfonts}
\usepackage{graphicx}

\newcommand{\dashedmidrule}{\noalign{\vskip\aboverulesep}\hdashline\noalign{\vskip\belowrulesep}}

\def\paperID{18964} 
\def\confName{CVPR}
\def\confYear{2026}

\title{
    \begin{minipage}{0.12\textwidth} 
        \raggedleft
        \includegraphics[height=1.5cm]{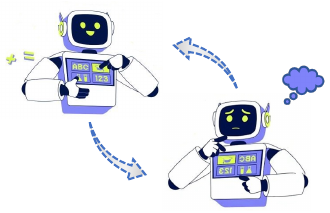} 
    \end{minipage}%
    \hspace{0.005cm} 
    \begin{minipage}{0.8\textwidth} 
        \centering
        \textbf{EvoLMM: Self-Evolving Large Multimodal Models with Continuous Rewards}
    \end{minipage}
}

\author{\\ {Omkar Thawakar}\textsuperscript{1$\dagger$} \quad{Shravan Venkatraman}\textsuperscript{1$\dagger$} \quad{Ritesh Thawkar}\textsuperscript{1$\dagger$}\quad {Abdelrahman M Shaker}\textsuperscript{1} \\
    {Hisham Cholakkal}\textsuperscript{1} \quad  {Rao Muhammad Anwer}\textsuperscript{1,2} \quad  
     {Salman Khan}\textsuperscript{1,3} \quad  {Fahad Khan}\textsuperscript{1,4}\\
     \fontsize{11pt}{12pt}\selectfont \textsuperscript{1}Mohamed bin Zayed University of AI,  \textsuperscript{2}Aalto University, \\
     \textsuperscript{3}Australian National University,
     \textsuperscript{4}Linköping University \\
     \fontsize{10pt}{12pt}\selectfont \faEnvelopeO \hspace{2pt} \{{omkar.thawakar, shravan.venkatraman, ritesh.thawkar}\}@mbzuai.ac.ae \\
 {\hypersetup{urlcolor=blue}
\fontsize{11pt}{12pt}\selectfont \faExternalLink \hspace{2pt} \href{https://mbzuai-oryx.github.io/EvoLMM/}{https://mbzuai-oryx.github.io/EvoLMM/}}
}

\begin{document}
\maketitle
\input{sec/0_abstract}    
\input{sec/1_intro}
\input{sec/2_related_work}

\input{sec/3_method}

\input{sec/4_experiments}
\input{sec/5_conclusion}

{
    \small
    \bibliographystyle{ieeenat_fullname}
    \bibliography{main}
}
\input{sec/X_suppl}



\end{document}

%% file: sec/0_abstract.tex
\begin{abstract}
Recent advances in large multimodal models (LMMs) have enabled impressive reasoning and perception abilities, yet most existing training pipelines still depend on human-curated data or externally verified reward models, limiting their autonomy and scalability. In this work, we strive to improve LMM reasoning capabilities in a purely unsupervised fashion (without any annotated data or reward distillation). To this end, we propose a self-evolving framework, named EvoLMM, that instantiates two cooperative agents from a single backbone model: a Proposer, which generates diverse, image-grounded questions, and a Solver, which solves them through internal consistency, where learning proceeds through a continuous self-rewarding process. This dynamic feedback encourages both the generation of informative queries and the refinement of structured reasoning without relying on ground-truth or human judgments. When using the popular Qwen2.5-VL as the base model, our EvoLMM yields consistent gains upto $\sim$3\% on multimodal math-reasoning benchmarks, including ChartQA, MathVista, and MathVision, using only raw training images. 
We hope our simple yet effective approach will serve as a solid baseline easing future research in self-improving LMMs in a fully-unsupervised fashion. Our code and models are available at \url{https://github.com/mbzuai-oryx/EvoLMM}. 
\def\thefootnote{$\dagger$}\footnotetext{Equal contribution.}

\end{abstract}

%% file: sec/1_intro.tex
\vspace{-1em}
\section{Introduction}
\label{sec:intro}

\begin{figure}[t] 
    \centering
    \includegraphics[width=0.95\linewidth]{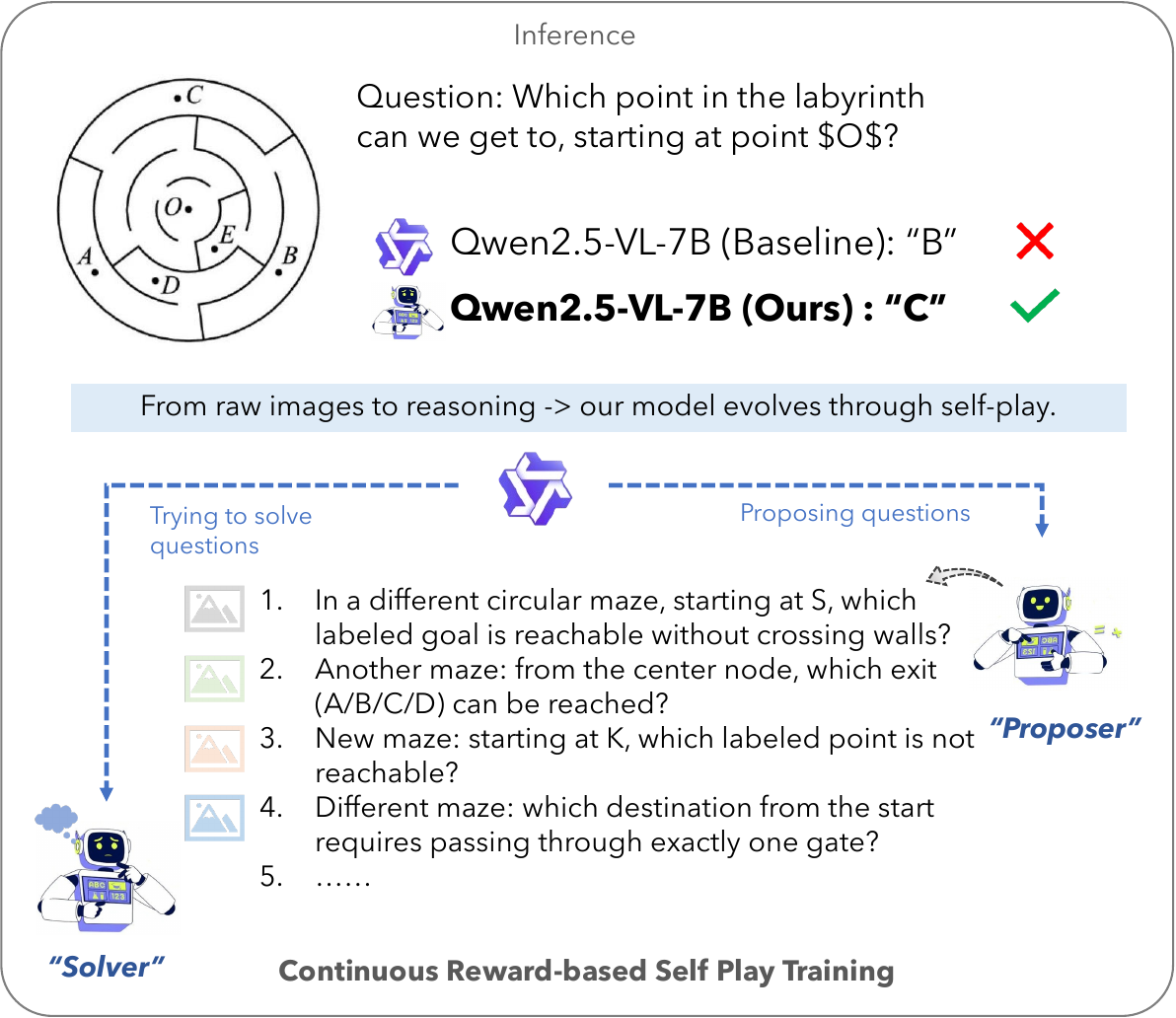}
    \vspace{-0.5em}
    \caption{
    \textbf{Illustration of our fully unsupervised self-evolving LMM framework (EvoLMM).} 
    Our EvoLMM enables a base LMM to improve its reasoning ability \emph{without any human labels, metadata, or external reward models}. 
    Given only raw images, a \emph{proposer} first generates visually grounded questions, and a \emph{solver} attempts to answer them multiple times. 
    The degree of agreement among solver responses produces a \emph{continuous self-consistency reward}, forming a closed-loop training signal that drives both modules to co-evolve. 
    }
    \label{fig:teaser}
    \vspace{-2em}
\end{figure}

Large multimodal models (LMMs) that jointly handle vision and language have made substantial progress in recent years. They excel at tasks like image captioning~\cite{Lu2025OmniCaptionerOCA,Jung2025VisualANA,Peng2025MattersTFA,Li2025DescribeEarthDAA}, visual question answering~\cite{Yang2025MAGICVQAMAA,Jiang2025FastOSA,Zhou2025HierarchicalVRA,Peng2025MVCoReMVA}, and multimodal reasoning~\cite{Yang2025R1OnevisionAGA,Zhang2025R1VLLTA,Wang2025VisualPRMAEA,Peng2025SkyworkRPA,thawakar2025llamav} by leveraging large-scale pre-training and fine-tuning on annotated datasets. However, despite these advances, two fundamental dependencies remain: (i) the dependence on human-curated annotations or metadata~\cite{Hudson2019GQAANA,Zhao2023MMICLEVA,AlTahan2024UniBenchVRA,Dai2024HumanVLMFFA,Liu2024MMDUAMA}, and (ii) the dependency on externally defined reward models or evaluators (e.g., human-labeled correctness, knowledge-distilled reward networks)~\cite{wang2025vision,cai2025llavakdframeworkdistillingmultimodal,shu2024llavamod,11094175,li2024self}. These dependencies pose scalability limitations, hinder domain generalization, and restrict deployment in settings where annotations or reward models are unavailable.


Recent research in large language models (LLMs) explores the idea of self-evolving models that not only solve tasks but also generate the tasks they learn from. For instance, few works~\cite{chen2025self,chen2025multi} aim to improve base LLM by generating its own questions and attempting to answer them, using the agreement among multiple sampled answers as a self-reward signal. Such a technique can be further improved~\cite{chen2025multi} by splitting the model into interacting roles such as \emph{Proposer}, \emph{Solver}, and Judge that co-exist and train themselves through reinforcement learning. This research direction is pushed further \cite{zhang2025darwin} by incorporating open-ended scheme where the system incorporates an open-ended evolution to continually refine internal behaviors. However, these research efforts are largely constrained to the \textit{language-only} domain, where reasoning is explicit and correctness can be inferred through textual consistency or symbolic execution. Moreover, these approaches still either rely on implicit evaluators (e.g., external Judge module, execution feedback) or explicit evaluators, implying that the learning signal is not yet fully free of external priors.

In the multimodal (LMM) domain, 
few recent works aim to improve LMMs through self-rewarding training signals, thereby reducing human supervision. For instance, the work of ~\cite{li2025self} proposes a self-rewarding reasoning strategy where the model decomposes each task into a perception step and a reasoning step, and then evaluates with internal bootstrapping. Similarly, ViPER~\cite{zhang2025viper} introduces a closed-loop framework that gradually strengthens visual perception ability through image-level and instance-level reconstruction, combined with reinforcement style updates driven by internally synthesized data. While these approaches represent progress in self-improving LMMs, they still rely on structured intermediate signals such as reconstruction objectives, heuristic quality filtering, or semantic similarity-based scoring. Consequently, their self-evolution remains partially guided by external priors rather than emerging purely from the model’s own reasoning dynamics.

In this work, we investigate a more fundamental form of autonomy:
\emph{Can a large multimodal model (LMM) enhance its own reasoning capability without relying on human-annotated supervision or any external reward signals?}
To this end, we propose a purely unsupervised self-evolving framework, named EvoLMM, that learns entirely through internal consistency (see Figure.~\ref{fig:teaser}).
Our approach instantiates two cooperative roles from a single backbone model: a \emph{Proposer}, which generates diverse and visually grounded mathematical questions from unlabeled images, and a \emph{Solver}, which attempts to solve these questions. The framework evolves through a continuous self-reward mechanism, where the \emph{Solver} is rewarded by the fraction of consistent or correct answers within its own generations. It is worth mentioning that no ground-truth labels, metadata, or external reward models are used at any stage.
This simple yet effective design enables the model to construct, evaluate, and refine its own training signal directly from raw multimodal inputs. 
The framework naturally scales to large models and domains where ground-truth supervision is unavailable or expensive.
Despite training exclusively on raw images, we show consistent improvements across multiple multimodal math reasoning benchmarks over the base model. Further, our qualitative analysis reveals that the \emph{Proposer} learns to generate increasingly complex visual problems, while the \emph{Solver} develops more structured reasoning chains over time, suggesting emergent self-evolution.

\noindent In summary, our main contributions are:
\begin{itemize}
    \item We introduce a self-evolving multimodal framework, named EvoLMM, that enables a base LMM to improve without human labels, metadata, or external reward models. The framework decomposes the model into two internal roles, \emph{Proposer} and \emph{Solver}, forming a closed-loop propose-solve cycle trained solely through internal consistency feedback. 
    \item We develop a continuous self-rewarding mechanism based on multi-sample answer consistency, which replaces both learned discrete reward models and semantic similarity scoring used in prior LMM self-evolution approaches. This continuous internal reward signal provides smooth gradients and stable optimization, enabling consistent improvement in performance. 
    \item We empirically validate EvoLMM on mathematical visual reasoning benchmarks, with absolute gains of $\sim2–3\%$ over the Qwen-2.5-VL-7B baseline using only raw images during training. We further analyze the evolution of our propose-solve mechanism where the difficulty level gradually progresses and maintains stable learning, showing that the model naturally develops more structured and grounded reasoning behaviors over time.
    Furthermore, we show that internal consistency can serve as a viable supervision signal for open-ended multimodal learning.
\end{itemize}






%% file: sec/2_related_work.tex
\section{Related Work}
\label{sec:related_works}
In the context of large language models (LLMs), several recent works \cite{chen2025self,shafayat2025can,chen2025multi} aim to self-improve the model capabilities (e.g., scientific reasoning) without significant human annotations. The work of \cite{chen2025self} proposes an approach to improve the implicit reasoning capabilities of the base model through self-play and majority-vote reward signals. Sheikh \citet{shafayat2025can} explores self-consistency as a training reward within an reinforcement learning (RL) framework, demonstrating performance on synthetic reasoning task. Recently, the work of \cite{chen2025multi} introduces a self-play based method where three roles (\emph{Proposer}, \emph{Solver}, and \emph{Judge}) are instantiated from the same base LLM and trained via RL to construct questions, answers, and the evaluation inter-play between them. 
While achieving promising results, these approaches assume a certain form of verifiable environment in the form of a judge \cite{chen2025multi} or heuristic and discrete rewards (e.g., majority vote) \cite{chen2025self, shafayat2025can} which can likely struggle with model collapse. 

In the context of large multimodal models (LMMs), \citet{li2025self} introduces a self-rewarding technique to improve visual understanding by decomposing it into perception and language-reasoning via explicit perception reward by the LMM itself. However, the self-reward mechanism relies on the SFT cold-start dataset to learn the desired format. The work of \citet{zhang2025viper} leverages image-level and instance-level reconstructions in a two-stage RL loop to iteratively improve perceptual capabilities. However, this image-and instance-level reconstruction of meta-data relies on an external model (e.g., OmniGen2~\cite{wu2025omnigen2} and Qwen-Image~\cite{wu2025qwen}). Additionally, a recent work Vision-Zero~\cite{wang2025vision}, trains LMMs through a multi-agent social deduction game (“Who is the Spy?”) played directly on images. The framework introduces adversarial roles (spy vs.\ civilians), iterative clue and voting rounds, and alternates between self-play and verifiable reinforcement learning (RLVR) depending on whether the identity of the spy is known. Although Vision-Zero does not require explicit human annotations, it relies on constructing image pairs with subtle, controlled differences produced through automated rendering, image editing, or regeneration pipelines relying on external generative systems (GPT-4o~\cite{gpt4o} and Gemini~\cite{comanici2025gemini}).



\noindent \textbf{Our Approach.} 
Different to aforementioned approaches, our approach aims to self-improve the multimodal capabilities (e.g., multimodal math reasoning) of a base LMM \textit{without} any annotations and external reward-model based self-evolution loop. Specifically, our framework uses only raw images (without metadata, bounding-boxes, and human labels) and employs a \emph{Proposer–Solver} pair where (i) the \emph{Proposer} generates questions from images and (ii) the \emph{Solver} attempts answers over multiple attempts, where the reward signal is the fraction of correct answers among the model's own outputs, effectively scaling with the degree of answer agreement. In this way, the proposed scheme avoids external judge, human annotated datasets (e.g., SFT cold-start), knowledge-distillation, and discrete rewards. By leveraging a \textit{continuous reward} design, our training process systematically induces a curriculum, where the \emph{Proposer} gradually explores questions that are neither trivial nor impractical, thereby stabilizing the \emph{Solver} learning and mitigating the \emph{Proposer} to degenerate to zero-rewards.

%% file: sec/3_method.tex
\section{Method}

\paragraph{Problem Formulation.}
We consider the \emph{unsupervised self-evolving} vision--language (multimodal) reasoning setting, where no question--answer annotations, metadata, or external reward models are available. Let $\mathcal{X}=\{x\}$ denote an unlabeled collection of images. Our goal is to enable a single pre-trained base large multimodal model (LMM) improve itself by generating its own training signal directly from these images. To this end, we instantiate two policies:

\begin{itemize}
    \item \textbf{\emph{Proposer}} $\pi_{\phi}(q\,|\,x)$ that produces a visually grounded question $q$ for image $x$, and
    \item \textbf{\emph{Solver}} $\pi_{\theta}(y\,|\,x,q)$ that attempts to answer $q$.
\end{itemize}

\noindent For each proposed question, we draw $N$ independent answer samples $y_{1:N} \sim \pi_{\theta}(\cdot\,|\,x,q)$ and compute the empirical answer distribution $p(a\,|\,x,q)$. Let $\hat{a}=\arg\max_a p(a\,|\,x,q)$ denote the majority answer, and let
\begin{equation}
H(x,q) = -\sum_a p(a\,|\,x,q)\log p(a\,|\,x,q)
\end{equation}
be the \emph{Solver}-consensus entropy. This quantity serves as a proxy for question difficulty: trivial questions lead to low entropy, whereas ambiguous or poorly grounded questions produce high entropy.

 \begin{figure*}[t!]
\includegraphics[width=\textwidth]{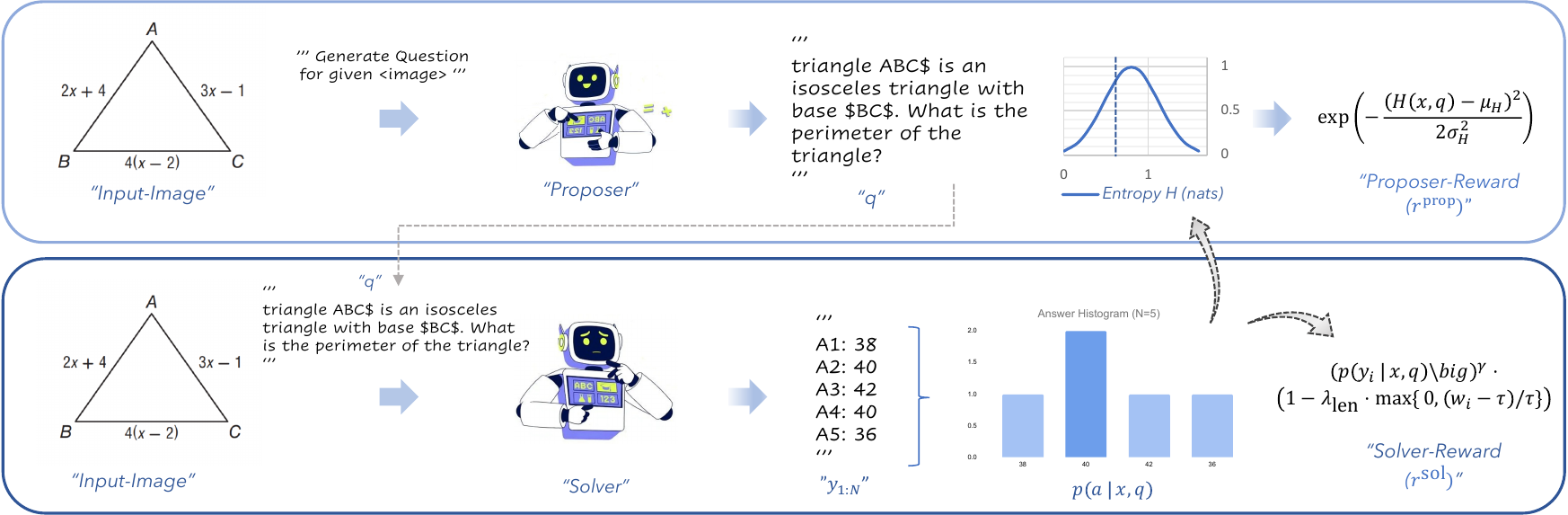}
\caption{
\textbf{Overview of our \emph{Proposer}–\emph{Solver} based self-evolving framework (EvoLMM).}
Given only a raw visual input (e.g., multimodal chart), the \emph{Proposer} module generates a question $q$ about the image content.
The \emph{Solver} then produces multiple answer samples $y_{1:N}$, forming an empirical answer distribution $p(a\,|\,x,q)$.
The \emph{Solver reward} $r^{\text{sol}}$ is a continuous, self-supervised signal based on the likelihood of each answer sample, modulated by a length penalty that constrains the \emph{Solver's} response format.
The \emph{Proposer reward} $r^{\text{prop}}$ is an entropy-based band-pass function that encourages moderate difficulty questions where \emph{Solver} is not completely correct and certain. By rewarding this moderate-entropy window, the \emph{Proposer} gradually learns to generate questions that are challenging enough to stimulate reasoning while remaining solvable, forming an automatic curriculum without external supervision. 
Both modules are optimized with standard REINFORCE objectives 
regularized by token-level KL constraints to reference policies.
This closed-loop training enables jointly refining question generation and reasoning \emph{using only images}, without any annotated Q\&A pairs, discrete rewards, or external verifiers.
}
\label{fig:architecture}
\end{figure*}

\subsection{Unsupervised Self-Evolving LMM}

In this work, we design an unsupervised self-evolving LMM approach (EvoLMM) motivated by the recent self-question-based technique (SQLM)~\cite{chen2025self} in the LLM domain. The SQLM improves an LLM by generating its own questions and reinforcing answer agreement. However, SQLM relies on a \emph{discrete} majority-vote reward, which we empirically show to be sub-optimal in the multimodal setting where the LMM is required to ground its questions based on the input image and perform complex multimodal reasoning (e.g., visual math reasoning). We observe that early-stage \emph{Solver} outputs over images are highly variable, leading to frequent zero-reward updates and unstable optimization. This prevents the model from making gradual progress when using discrete majority-vote reward. 



In our EvoLMM approach, we replace the discrete reward with a \emph{continuous self-consistency} signal that scales with the degree of answer-agreement, as shown in Figure~\ref{fig:architecture}. The continuous self-consistency signal provides meaningful gradient feedback even when visual reasoning is only partially correct. 
We observe this to stabilize the learning in the presence of visual uncertainty that requires domain knowledge. 
Our continuous self-consistency reward enables the \emph{Proposer} and \emph{Solver} to co-evolve smoothly, forming a curriculum in which question difficulty and reasoning quality improve gradually using only raw images.

\begin{figure*}[t]
    \centering
    \includegraphics[width=\textwidth]{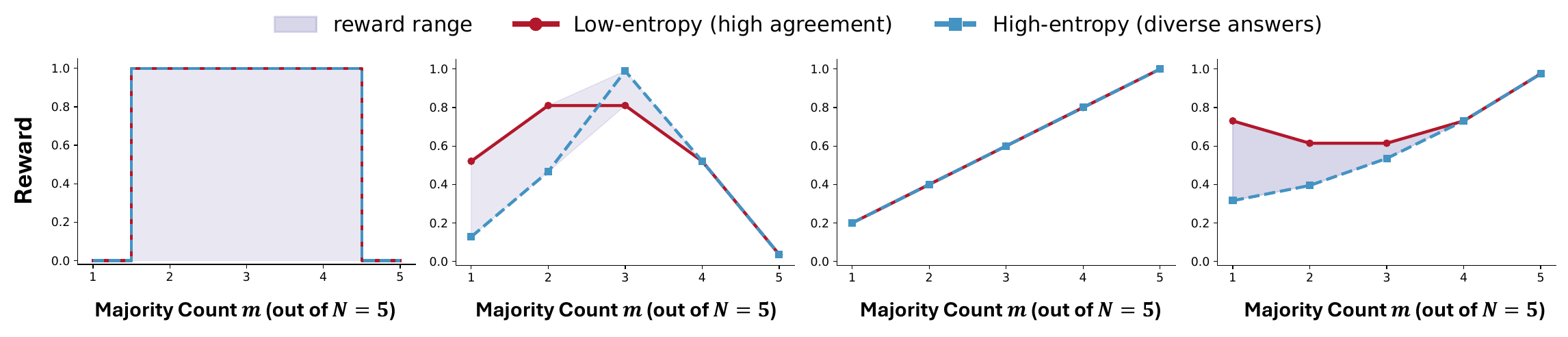}
    {\footnotesize
    \noindent
    \begin{minipage}[t]{.245\textwidth}\centering
        \hspace{4.0em}(a) Proposer (Discrete)
    \end{minipage}\hfill
    \begin{minipage}[t]{.245\textwidth}\centering
        \hspace{3.5em}(b) Proposer (Continuous)
    \end{minipage}\hfill
    \begin{minipage}[t]{.245\textwidth}\centering
        \hspace{2.3em}(c) Solver (Discrete)
    \end{minipage}\hfill
    \begin{minipage}[t]{.245\textwidth}\centering
        \hspace{1.0em}(d) Solver (Continuous)
    \end{minipage}
    }
    \caption{
      \textbf{Comparison of \emph{Proposer} and \emph{Solver} rewards under discrete and our continuous formulations for a single iteration.}
      Each panel shows rewards for \emph{low-entropy} (high-agreement) and \emph{high-entropy} (diverse-answer) cases for a single iteration for number of \emph{Solver} responses \(N{=}5\). In case of \emph{Proposer} (left), discrete rewards  collapse them into identical plateaus, providing weaker learning signals. In contrast, our continuous reward varies smoothly and distinguishes \emph{Solver} response patterns. In case of \emph{Solver} (right), the discrete reward increases linearly only with majority count and does not reflect partial progress, leading to sparse and unstable learning signals during early training. Instead, our \emph{continuous Solver reward} scales smoothly with agreement, enabling the \emph{Solver} to improve reasoning consistency gradually. Collectively, our continuous reward formulation creates a more stable self-evolution loop, where the \emph{Proposer} and \emph{Solver} co-adapt toward more grounded and consistent reasoning behavior.
    }
    \label{fig:analysis_plots}
\end{figure*}

\begin{figure}[t]
    \centering
    \includegraphics[width=\linewidth]{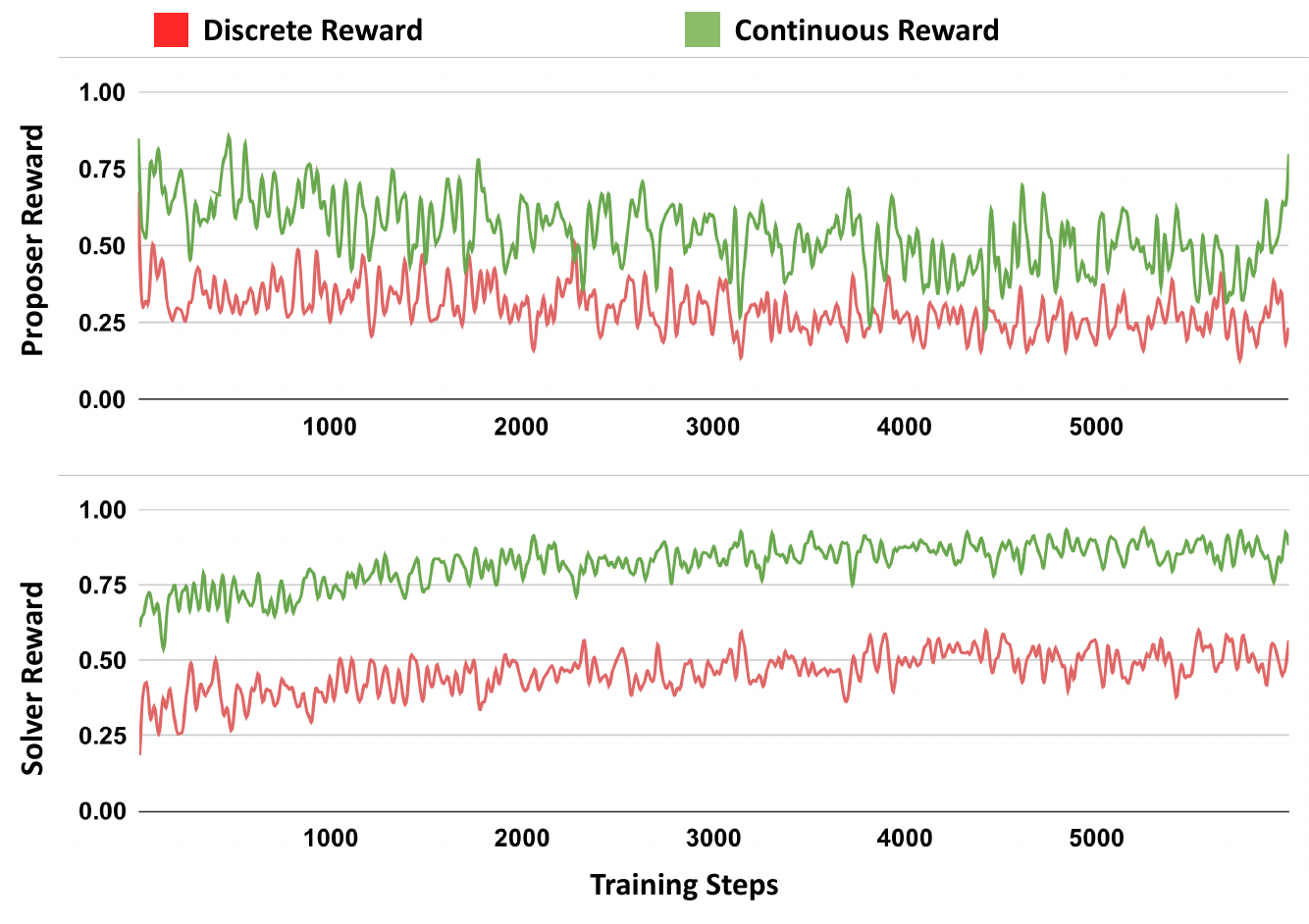}
    \caption{
        \textbf{Comparison between the discrete vs. our continuous reward progression during training.} Top: The discrete majority-vote reward (red) remains low and unstable during training, providing a weak learning signal due to \emph{Solver} output variability in early stages. In contrast, our continuous self-consistency reward (green) produces stable and valuable feedback, enabling the \emph{Proposer} to consistently generate moderate difficulty, informative questions. Bottom: Guided by our \emph{Proposer} with continuous reward, we observe the \emph{Solver} (green) to be more consistent and stable compared to its discrete (red) counterpart. 
    }
    \label{fig:solver_graphs}
\end{figure}

\noindent \textbf{Continuous Self-Consistency Reward.}
We reward the \emph{Solver} using a continuous self-supervised signal that scales smoothly with self-consistency and gently penalizes verbosity. For each sample $y_i$, we define
\begin{equation}
r^{\text{sol}}_i = \big(p(y_i\,|\,x,q)\big)^{\gamma}\cdot\Big(1 - \lambda_{\text{len}}\cdot\max\{0,(w_i-\tau)/\tau\}\Big)
\label{eq:solver_reward}
\end{equation}
where $p(y_i\,|\,x,q)$ is the \emph{Solver's} own agreement score, $\gamma\!\in\!(0,1]$ controls reward softness (lower values accentuate mid-confidence differences), $w_i$ counts the number of words preceding the \texttt{<answer>} tag, and $\tau$ is target brevity threshold. This yields dense, bounded rewards that preserve gradient flow even without clear majority, encouraging both \emph{stable reasoning} (high agreement among outputs) and \emph{concise format}, as shown in Figure~\ref{fig:solver_graphs}. Unlike discrete majority-vote rewards, it provides a continuous signal that better separates mid-probability outcomes (see Figure~\ref{fig:analysis_plots}).

\noindent \textbf{Entropy-Guided Continuous \emph{Proposer} Reward.}
To avoid trivial or unsolvable questions, the \emph{Proposer} receives a smooth band-pass reward based on the \emph{Solver’s} answer entropy:
\begin{equation}
r^{\text{prop}} = \exp\!\left(-\frac{(H(x,q)-\mu_H)^2}{2\sigma_H^2}\right)
\end{equation}
which peaks when the \emph{Solver’s} uncertainty lies within a moderate range—neither overly certain ($H\!\approx\!0$, trivial or leading) nor overly uncertain (ambiguous or unsolvable), as shown in Figure~\ref{fig:analysis_plots}. This encourages the \emph{Proposer} to operate near the \emph{Solver’s} decision boundary, inducing an adaptive curriculum: as the \emph{Solver} improves, the \emph{Proposer} must generate slightly harder yet still solvable questions to remain within the entropy band (see Figure~\ref{fig:question_difficulty}).




\noindent \textbf{Training Optimization.}
In our framework, the \emph{Proposer} and \emph{Solver} are trained jointly in an online reinforcement-learning loop. Since no ground-truth labels or external scoring functions are available, both policies rely entirely on internally generated reward signals. To ensure stable learning, we adopt a REINFORCE policy gradient \cite{williams1992simple} with two key components: (i) moving-average baselines to reduce variance, and (ii) token-level KL regularization to monitor deviation from the pretrained base LMM.

Both policies share the same optimization form,
\begin{equation}
\label{eq:generic-loss}
\mathcal{L}_A = -\,\mathbb{E}\!\left[(r_A - b_A)\,\frac{1}{T_A}\sum_{t=1}^{T_A}\log \pi_A(y_t \mid h_t)\right] + \beta_A\,\overline{\mathrm{KL}}_A,
\end{equation}
where $A\!\in\!\{\text{\emph{Solver}},\text{\emph{Proposer}}\}$, $h_t$ denotes the text–image context up to token $t$ whereas, 
\begin{equation}
\overline{\mathrm{KL}}_A = \tfrac{1}{T_A}\sum_t \mathrm{KL}\!\big(\pi_A(\cdot\mid h_t)\,\|\,\pi_A^{\text{ref}}(\cdot\mid h_t)\big)
\label{eq:kl_divergence}
\end{equation}
 measures per-token divergence from a frozen reference model. 

\noindent Each policy employs an exponential moving-average baseline $b_A$ for variance reduction and a dynamic \emph{KL controller} that adjusts $\beta_A$ to maintain a target divergence budget:
\begin{equation}
\label{eq:beta-adapt}
\beta_A \leftarrow \operatorname{clip}\!\Big(\beta_A \cdot \exp\!\big(\eta \,\tfrac{\overline{\mathrm{KL}}_A-\tau_A}{\tau_A}\big),\;\beta_{\min},\beta_{\max}\Big).
\end{equation}
For the \emph{Solver}, this controller stabilizes the learning while allowing gradual adaptation towards better visual reasoning and consistent answer generation. For the \emph{Proposer}, the KL term is not tightly constrained: as it explores new question spaces, its divergence from the backbone (base LMM) increases, reflecting the emergence of more challenging queries. Please refer to Algorithm~\ref{alg:evolmm} for our end-to-end training optimization process.

\begin{figure}[t!]
    \includegraphics[width=\linewidth]{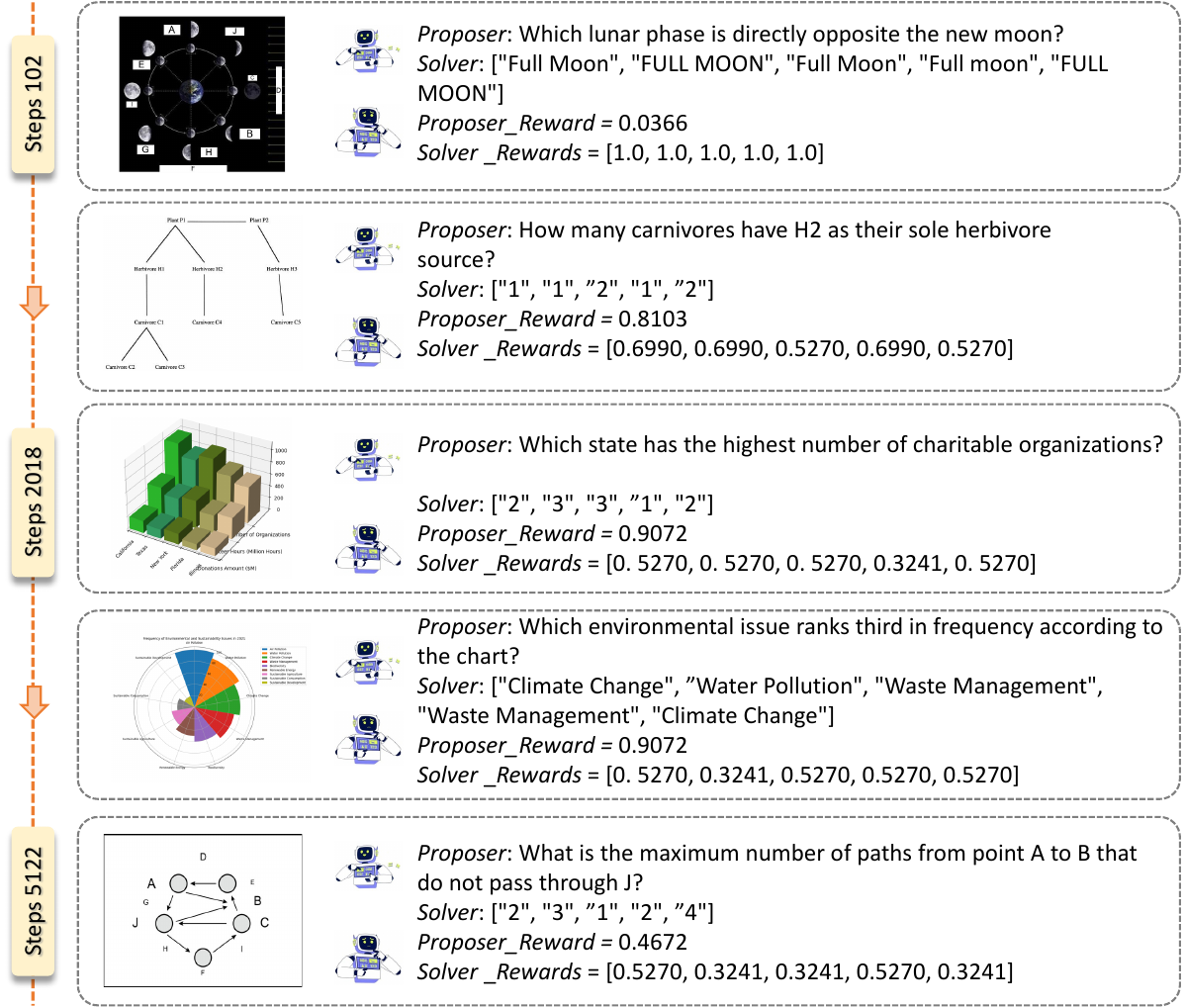}
    \caption{\textbf{Example showing progression of our continuous reward-based \emph{Proposer} questions generation along with rewards.}  From top (step 102) to bottom (step 5122), \emph{Proposer} increases question complexity, which in turn enhances the reasoning capabilities of \emph{Solver}. Refer suppl. material for more examples. }
\label{fig:question_evolution}
\end{figure}

\subsection{EvoLMM Training Insights}

\begin{algorithm}[t]
\caption{EvoLMM: Self-Evolving Training strategy with continious rewards.}
\label{alg:evolmm}
\KwIn{Unlabeled images $\mathcal{X}$; pretrained LMM backbone $\pi_{\text{ref}}$}
\KwOut{Trained Proposer $\pi_\phi$ and Solver $\pi_\theta$}
\tcp{Initialize Proposer, Solver}
Initialize Proposer $\pi_\phi \leftarrow$ LoRA($\pi_{\text{ref}}$) \\
Initialize Solver $\pi_\theta \leftarrow$ LoRA($\pi_{\text{ref}}$) \\
Initialize EMA baselines $b_{\text{prop}}, b_{\text{sol}}$ \\
Set number of answer samples $N$, entropy targets $(\mu_H, \sigma_H)$

\For{each training step}{
    Sample image $x \sim \mathcal{X}$

    \tcp{Step 1: Proposer generates a question}
    Generate question $q \sim \pi_\phi(q|x)$

    \tcp{Step 2: Solver answers the question multiple times}
    Sample $N$ answers $\{y_i\}_{i=1}^N \sim \pi_\theta(\cdot|x,q)$

    \tcp{Step 3: Compute empirical answer distribution}
    Estimate $p(a|x,q)$ and entropy $H(x,q)$

    \tcp{Step 4: Continuous Solver Reward}
    \For{$i=1$ {\bf to} $N$}{
        Compute agreement score $p(y_i|x,q)$ \\
        Compute reward:
        $$ r^{\text{sol}}_i = \left[p(y_i|x,q)\right]^\gamma
        \cdot \left(1-\lambda_{\text{len}}\max\{0,(w_i-\tau)/\tau\}\right) $$
    }

    \tcp{Step 5: Entropy-Based Proposer Reward}
    $$ r^{\text{prop}} = \exp\left(
        -\frac{(H(x,q)-\mu_H)^2}{2\sigma_H^2}
    \right) $$

    \tcp{Step 6: Policy Gradient Updates}
    Update Solver using REINFORCE:
    $$ \nabla_\theta \sim (r^{\text{sol}}_i - b_{\text{sol}})\nabla \log\pi_\theta $$

    Update Proposer every $K$ steps:
    $$ \nabla_\phi \sim (r^{\text{prop}} - b_{\text{prop}})\nabla \log\pi_\phi $$

    Apply KL-regularization controller:
    $$ \beta \leftarrow \beta \cdot \exp\left(\eta \frac{\mathrm{KL} - \tau}{\tau}\right) $$

}
\end{algorithm}

As discussed earlier, our \emph{Proposer}-\emph{Solver} training loop produces a joint self-evolving learning dynamics in which both question generation and answer reasoning gradually improve over the time using only images. In the absence of human annotations or external reward models, our model learns to rely entirely on the structure of its own outputs to guide the learning. 

To present further insights into our continuous self-consistency reward-based training, we conduct an experiment to post-train Qwen2.5-VL-7B \cite{qwen25vlreport} on ChartQA~\cite{chartqa} training images (without any question-answer pairs or metadata). As shown in Figure~\ref{fig:question_evolution}, our self-consistency reward yields a \emph{graded} score proportional to the empirical agreement among \emph{Solver} outputs. Even partial agreement (e.g., 2 out of 5 consistent samples) produces a meaningful positive reward. Furthermore, our continuous feedback reduces collapse and accelerates \emph{Solver} improvement (see Figure~\ref{fig:solver_graphs}).
The continuous \emph{Solver} reward changes smoothly and slightly dips at partial consensus, while the \emph{Proposer} peaks at moderate difficulty (see step 5122 in Figure~\ref{fig:question_evolution}).

In summary, our proposed continuous self-evolving design provides a non-zero learning signal even when the model is uncertain, avoiding stagnation observed in the discrete-reward self-questioning scheme. Further, our design enables the \emph{Proposer} to continuously adjust question difficulty to match the \emph{Solver’s} evolving capability, thereby mitigating model collapse.


%% file: sec/4_experiments.tex
\section{Experiments}
\label{sec:experiments}

\begin{figure}[t] 
    \centering
    \includegraphics[width=\linewidth]{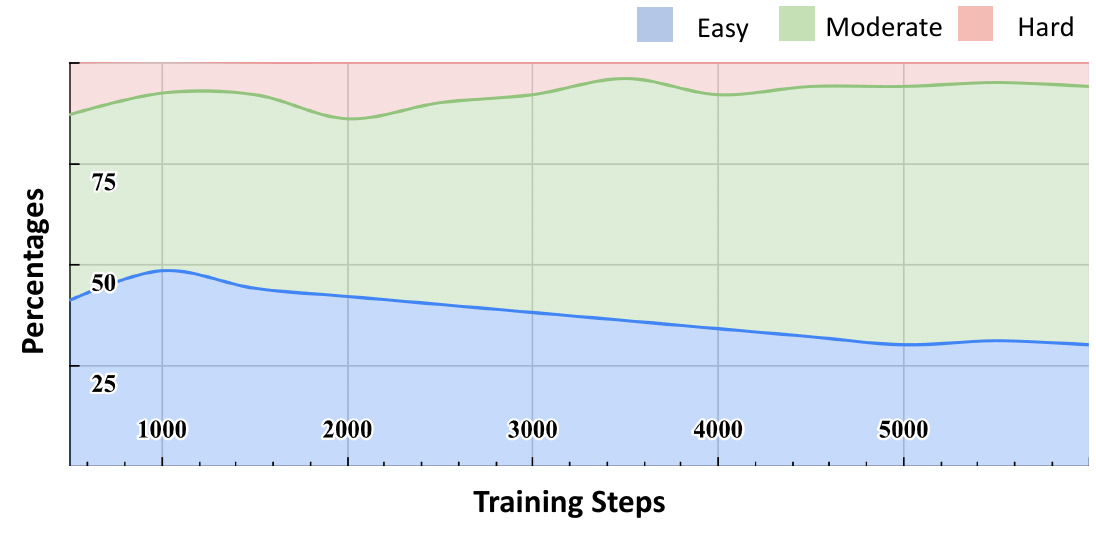}
    \caption{
    \textbf{Evolution of proposer question difficulty over training}.
    As self-evolution progresses, the proposer shifts from mostly easy or overly hard questions to a stable distribution dominated by moderate-difficulty ones. This reflects the emergence of an implicit curriculum: mid-entropy questions provide the most informative signal for the solver, leading to more stable and effective learning.
    }
    \label{fig:question_difficulty}
\end{figure}

\begin{table*}[t!]
\input{Tables/main_results}
\label{tab:main_results}
\end{table*}

\begin{table*}[t]
\input{Tables/train_settings}
\label{tab:train_setting}
\end{table*}

\subsection{Dataset}
We train and evaluate our method in the mathematical and visually grounded reasoning domain, which naturally requires integrating perception with structured, multi-step reasoning making it a strong testbed for self-evolving multimodal learning. These datasets contain rich visual patterns (trends, proportions, spatial relations, geometric structures) that allow models to generate meaningful questions directly from images without human annotations.


Our training uses only raw images, with no question–answer pairs or metadata. We sample roughly 1,000 images each from six widely used benchmarks such as ChartQA \cite{chartqa}, AI2D \cite{ai2d}, InfographicVQA \cite{infovqa}, PlotQA \cite{plotqa}, ChartX \cite{chartx}, and Geometry3K \cite{geovqa} totaling $\sim$6k images. These datasets cover charts, plots, scientific diagrams, and geometric figures, providing diverse visual contexts for the \emph{Proposer} to generate grounded mathematical questions. This diversity encourages the \emph{Solver} to develop stable, structured reasoning as the system self-evolves using internal consistency alone.

\subsection{Training Details}
We fine-tune the pretrained Qwen-2.5-VL-7B model using two lightweight LoRA adapters for the \emph{Proposer} and \emph{Solver} roles, keeping the backbone frozen to retain general multimodal capability. Training runs on a single node with 8$\times$ AMD MI250X GPUs using bfloat16 precision.
The solver is optimized with our continuous self-consistency reward computed from $N=5$ sampled answers, while the proposer receives an entropy-based band-pass reward encouraging mid-difficulty questions. Both policies are trained using KL-regularized REINFORCE with adaptive KL control for stability. The proposer is updated every 5 iterations following \cite{chen2025self}, while the solver receives gradients at each step. We apply AdamW with a $1\text{e}{-6}$ learning rate, weight decay of $0.01$, and gradient clipping at $1.0$. The continuous rewards use a length penalty of $0.10$ toward a 6-word target, solver softness exponent $\gamma=0.7$, and a Gaussian entropy reward for the proposer centered at $\mu=0.90$ with $\sigma=0.35$ (more details and comparisons on the update schedule and reward design are provided in the suppl. material.). Training is performed for $6000$ steps with a batch size of $1$ using a constant learning rate schedule. 
No QA annotations, metadata, or external reward models are used at any stage.

\subsection{Evaluation}

We evaluate our self-evolving framework on a broad suite of mathematical and visually grounded reasoning benchmarks, including ChartQA \cite{chartqa}, MathVista \cite{mathvista}, MathVision \cite{mathvision}, MathVerse \cite{mathverse}, InfoGraphic-VQA \cite{infovqa}, AI2D \cite{ai2d}, ScienceQA \cite{scienceqa}, and MMMU \cite{mmmu}. These datasets test a range of capabilities from chart interpretation and symbolic math to diagram understanding—and all metrics follow their standard accuracy protocols.
We use the official evaluation splits and apply identical inference settings across models, ensuring that performance differences arise solely from our self-evolving training rather than task-specific tuning. All evaluations are run on AMD MI250X GPUs using the lmms-eval framework \cite{lmmseval}, with HuggingFace Transformers v4.38 and bfloat16 precision for consistency with training.



\begin{table*}[t]
\input{Tables/ablations}
\label{tab:ablations}
\end{table*}

\begin{table*}[t]
\input{Tables/model_size}
\label{tab:model_size}
\end{table*}

\subsection{Main Results}

In Table~\ref{tab:main_results}, we compare our proposed EvoLMM framework against two baselines:
(1) the original pretrained Qwen2.5-VL-7B model~\cite{qwen25vlreport} without any additional training, and
(2) a direct multimodal adaptation \cite{chen2025self} that uses a discrete majority-vote reward during training.
\textit{Qwen2.5-VL-7B (Baseline).}
The baseline model performs well on visually grounded reasoning tasks that require direct reading or surface-level interpretation (e.g., ChartQA trend identification, InfoGraphic-VQA object recognition). However, the model’s performance drops on tasks requiring multi-step inference or symbolic reasoning (e.g., MathVista, MathVerse), indicating that pretrained multimodal alignment alone is insufficient for deeper reasoning.
We also report results from recent {Vision-Zero}~\cite{wang2025vision}, whose self-play formulation bears partial similarity to ours through autonomous multi-agent learning. However, Vision-Zero leverages synthetic image pairs generated using external models such as GPT-4o \cite{gpt4o} and Gemini~\cite{comanici2025gemini}, introducing implicit supervision that falls outside our unsupervised setting. 
{Vision-SR1}~\cite{li2025self} differs entirely from our constraints, as it relies on paired question–answer supervision across both perception and reasoning stages.

\noindent \textbf{Discrete-Reward Adaptation.}
We first test a direct multimodal adaptation of the discrete self-questioning reward, where the \emph{Solver} is rewarded only when multiple sampled answers exactly match. In practice, the reward quickly becomes sparse as \emph{Solver} outputs remain highly diverse in early training, leading to unstable optimization. The discrete variant yields only marginal gains ($+0.3 - 0.6\%$) and occasionally degrades performance, confirming that majority-vote rewards are too brittle and unstable for multimodal reasoning.

\noindent \textbf{Continuous-Reward (Ours).}
Replacing the discrete signal with our continuous self-consistency reward provides a smoother and more informative gradient. Training progresses steadily even under uncertainty, leading to consistent improvements of +2–3\% across major benchmarks. For instance, accuracy rises from {68.4\% → 70.5\%} on \emph{MathVista} and {84.0\% → 86.7\%} on \emph{ChartQA}. Other than mathematical reasoning, our method also keep competitive improvement of +1.2\% on \emph{ScienceQA}. These results demonstrate that continuous internal rewards yield stable, annotation-free self-evolution, strengthening both reasoning reliability and answer consistency.

\subsection{Comparison Across Training Settings}
We evaluate the self-evolving framework under three fine-tuning strategies such as LoRA~\cite{lora}, QLoRA~\cite{qlora}, and full-parameter fine-tuning—while keeping the \emph{Proposer–Solver} pipeline and continuous reward formulation identical across settings (see Table~\ref{tab:train_settings}).

\noindent \textbf{LoRA (default).}
LoRA~\cite{lora} achieves the best trade-off between stability and adaptability. By updating only a small fraction of parameters, it preserves pretrained multimodal alignment while enabling effective \emph{Proposer–Solver} co-evolution. LoRA delivers consistent and stable gains, improving accuracy from {68.4\%$\rightarrow$70.5\%} on \emph{MathVista} and {84.0\%$\rightarrow$86.7\%} on \emph{ChartQA}, without any supervised data.

\noindent \textbf{QLoRA.}
The quantized variant QLoRA~\cite{qlora} (4-bit base + high-precision adapters) reduces memory usage by nearly $3\times$, making it suitable for limited hardware. However, quantization noise slightly weakens solver consistency, yielding smaller but still positive gains (e.g., {68.4\%$\rightarrow$68.92\%} on \emph{MathVista}). It indicates that high-precision activations are important when learning from fine-grained continuous rewards.

\noindent \textbf{Full Fine-Tuning.}
Full-parameter updates lead to faster initial learning but exhibit reduced stability in the self-evolving setting. Without explicit supervision, the \emph{Solver} tends to overfit early biases, and the \emph{Proposer} occasionally collapses into repetitive or low-diversity question modes. Moreover, because our reward formulation includes a KL regularization term that constrains divergence from the pretrained base model, full fine-tuning interacts unfavorably with this objective causing the model to drift away from its original alignment and degrade baseline performance. Consequently, full fine-tuning yields lower final accuracy than LoRA, despite higher compute cost and greater instability.


\subsection{Self Evolution Across Different Backbones}
To evaluate the generality of our approach, we apply the same \emph{Proposer–Solver} training setup to several large multimodal models, including \textit{Qwen3-VL-8B-Instruct}~\cite{qwen25vlreport}, \textit{InternVL3-8B}~\cite{zhu2025internvl3}, \textit{Gemma-3-12B-it}~\cite{team2025gemma}, and \textit{Llama-3.2-11B-Vision-Instruct}~\cite{meta2024llama3_2}. As shown in Table~\ref{tab:ablations}, in all cases, the models are trained without question–answer supervision, using only the proposed continuous self-consistency reward.
We observe consistent improvements of roughly {$+2-3\%$} across all backbones, confirming that our method is architecture-agnostic and does not rely on specific dataset alignments or handcrafted supervision. Models with stronger visual grounding (e.g., InternVL3-8B~\cite{zhu2025internvl3}) show faster convergence in early training, while those with stronger linguistic reasoning (e.g., Gemma-3-12B~\cite{team2025gemma}) display larger late-stage gains as solver reasoning stabilizes. These results demonstrate that continuous self-reward effectively generalizes across diverse LMM families. This shows that our reward mechanism enhances structured visual reasoning rather than surface-level pattern matching, providing a scalable and backbone-independent path toward autonomous multimodal reasoning.

\subsection{Scaling Behavior Across Model Sizes}
We evaluate our self-evolving framework across the Qwen2.5-VL family (7B \& 72B) using the same \emph{Proposer–Solver} setup and continuous reward, without modifying data or optimization settings (refer Table~\ref{tab:model_size}).
Consistent gains are observed at all scales, with the 7B model improving from $84.0\%$ → $86.7\%$ on ChartQA and $68.4\%$ → $70.5\%$ on MathVista, while larger variants (e.g., 72B) achieve stronger absolute performance $88.20\%$ → $91.04\%$ on ChartQA and $93.36\%$ → $94.63\%$. 
Overall, these results confirm our continuous self-consistency reward scales reliably across model sizes, enabling stable improvements without task-specific supervision or external verifiers.

%% file: Tables/main_results.tex
\centering
\setlength{\tabcolsep}{4pt}
\resizebox{\textwidth}{!}{%
\begin{tabular}{l|cccccccc}
\toprule
\rowcolor{gray!15} \textbf{Model} & \textbf{ChartQA} & \textbf{MathVista} & \textbf{MathVision} & \textbf{MathVerse} & \textbf{InfoGraphic-VQA$_{val}$} & \textbf{AI2D} & \textbf{ScienceQA} & \textbf{MMMU$_{val}$} \\
\midrule
Vision-Zero$^{\dagger}$ (CLEVR) \cite{wang2025vision} & $84.24$ & $68.43$ & $23.96$ & $43.86$ & $80.35$  & $82.64$ & $88.50$ & $51.44$ \\
\dashedmidrule
Qwen2.5-VL-7B (Baseline) \cite{qwen25vlreport} & $84.00$ & $68.46$ & $23.91$ & $43.78$ & $80.44$ & $82.61$ & $88.30$ & $51.11$ \\
Qwen2.5-VL-7B + Discrete Reward & $84.62$ & $68.88$ & $22.52$ & $42.10$ & $80.52$  & $82.18$ & $87.98$ & $50.84$ \\
\rowcolor{cyan!10}\textbf{Qwen2.5-VL-7B + Cont. Reward} & \textbf{$86.70$} & \textbf{$70.52$} & \textbf{$24.81$} & \textbf{$44.88$} & \textbf{$81.06$} & \textbf{$83.41$} & \textbf{$89.50$} & \textbf{$52.01$} \\
\rowcolor{cyan!10} 
$\Delta$ Improvement & $+2.7$\% & $+2.06$\% & $+0.9$\% & $+1.1$\% & $+0.62$\% & $+0.8$\% & $+1.2$\% & $+0.9$\% \\
\bottomrule
\end{tabular}
}
\vspace{-0.5em}
\caption{
\textbf{Evaluation results across eight multimodal mathematical and visual reasoning benchmarks.} 
We compare the pretrained Qwen2.5-VL-7B baseline, a discrete self-questioning adaptation, and our continuous self-evolving framework. The discrete reward provides little to no improvement, reflecting its instability in multimodal settings. In contrast, our continuous self-consistency reward yields consistent gains across all benchmarks, such as $84.00\%$ → $86.70\%$ on ChartQA and $88.30\%$ → $89.50\%$ on ScienceQA, demonstrating stable self-improvement from raw images without external supervision. Methods marked with an (${\dagger}$) uses external supervision.
}
\label{tab:vr_chartqa_summary}

%% file: Tables/train_settings.tex
\centering
\setlength{\tabcolsep}{4pt}
\resizebox{\textwidth}{!}{%
\begin{tabular}{l|cccccccc}
\toprule
\rowcolor{gray!15} \textbf{Model} & \textbf{ChartQA} & \textbf{MathVista} & \textbf{MathVision} & \textbf{MathVerse} & \textbf{InfoGraphic-VQA$_{val}$} & \textbf{AI2D} & \textbf{ScienceQA} & \textbf{MMMU$_{val}$} \\
\midrule
Qwen2.5-VL-7B (Baseline) & $84.00$ & $68.46$ & $23.91$ & $43.78$ & $80.44$ & $82.61$ & $88.30$ & $51.11$ \\

\textbf{Qwen2.5-VL-7B + LoRA} & $\textbf{86.70}$ & $\textbf{70.52}$ & $\textbf{24.81}$ & $\textbf{44.88}$ & $\textbf{81.06}$ & $\textbf{83.41}$ & $\textbf{89.50}$ & $\textbf{52.01}$ \\

Qwen2.5-VL-7B + QLoRA & $85.32$ & $68.92$ & $23.97$ & $43.82$ & $80.83$ & $82.75$ & $88.73$ & $51.71$ \\

Qwen2.5-VL-7B + Full-Finetune & $84.20$ & $68.41$ & $23.37$ & $43.77$ & $80.37$ & $82.64$ & $88.12$ & $51.23$ \\

\bottomrule
\end{tabular}
}
\vspace{-0.5em}
\caption{
\textbf{Comparison of our EvoLMM self-evolving framework under different parameter update strategies.} 
We evaluate LoRA, QLoRA, and full-parameter fine-tuning on Qwen2.5-VL-7B while keeping the same self-evolving training setup. LoRA achieves the strongest improvements, e.g., $84.00\%$ → $86.70\%$ on ChartQA and $88.30\%$ → $89.50\%$ on ScienceQA, demonstrating stable and effective adaptation while preserving pretrained multimodal grounding. QLoRA also improves performance but is slightly limited by quantization noise, while full fine-tuning yields minimal gains and in some cases regresses, likely due to overfitting without external supervision. These results highlight that parameter-efficient update schemes are better suited for continuous self-evolution without labeled data.
}
\label{tab:train_settings}

%% file: Tables/ablations.tex
\centering
\setlength{\tabcolsep}{4pt}
\resizebox{\textwidth}{!}{%
\begin{tabular}{l|cccccccc}
\toprule
\rowcolor{gray!15} \textbf{Model} & \textbf{ChartQA} & \textbf{MathVista} & \textbf{MathVision} & \textbf{MathVerse} & \textbf{InfoGraphic-VQA$_{val}$} & \textbf{AI2D} & \textbf{ScienceQA} & \textbf{MMMU$_{val}$}  \\
\midrule

Qwen2.5-VL-7B (Base) \cite{qwen25vlreport} & $84.00$ & $68.46$ & $23.91$ & $43.78$ & $80.44 $ & $82.61$ & $88.30$ & $51.11$ \\
\rowcolor{cyan!5}
\textbf{Qwen2.5-VL-7B (Ours)}  & $86.70$ & $70.52$ & $24.81$ & $44.88$ & $81.06$ & $83.41$ & $89.50$ & $52.01$ \\

\dashedmidrule
InternVL3-8B-Instruct  (Base) \cite{zhu2025internvl3} & $82.40$ & $65.2$ & $25.36$ & $31.62$ & $68.77$ & $83.19$ & $97.77$ & $52.78$ \\
\rowcolor{cyan!5}
\textbf{InternVL3-8B-Instruct (Ours)} & $84.97$ & $67.20$ & $26.44$ & $32.92$ & $69.39$ & $83.95$ & $98.13$ & $53.77$ \\

\dashedmidrule
Gemma3-12B-It (Base) \cite{team2025gemma} & $55.64$ & $60.13$ & $24.53$ & $28.96$ & $50.69$ & $79.05$ & $83.89$ & $48.11$ \\
\rowcolor{cyan!5}
\textbf{Gemma3-12B-It (Ours)} & $58.61$ & $62.13$ & $25.61$ & $30.26$ & $51.37$ & $79.85$ & $84.97$ & $49.10$ \\

\dashedmidrule
Llama-3.2-11B-Vision-Instruct (Base) \cite{meta2024llama3_2} & $29.24$ & $46.59$ & $23.47$ & $37.23$ & $56.69$ & $46.44$ & $56.87$ & $47.93$ \\
\rowcolor{cyan!5}
\textbf{Llama-3.2-11B-Vision-Instruct (Ours)} & $32.24$ & $48.59$ & $24.55$ & $38.53$ & $57.37$ & $47.32$ & $58.07$ & $48.92$ \\

\bottomrule
\end{tabular}
}
\vspace{-0.5em}
\caption{
\textbf{Effectiveness of our EvoLMM self-evolving framework across different large multimodal backbones.}
We apply the same \emph{Proposer--Solver} continuous self-consistency training to four LMM families without changing architecture, data, or supervision. On Qwen2.5-VL-7B, our method improves visual math reasoning performance consistently across all benchmarks (e.g., $84.00\%$ $\rightarrow$ $86.70\%$ on ChartQA and $88.30\%$ $\rightarrow$ $89.50\%$ on ScienceQA). We observe similar trends when transferring to, InternVL3-8B, and Llama-3.2-11B-Vision-Instruct: each model benefits from continuous self-evolution, typically gaining $+1\text{--}3$ accuracy points on chart/diagram reasoning tasks and $+1\text{--}2$ points on math and science QA. These improvements occur \emph{without} annotations, handcrafted curriculum, image editing, or external verifiers, demonstrating that our continuous reward mechanism is model-agnostic and transferable across diverse LMMs.
}
%

%% file: Tables/model_size.tex
\centering
\setlength{\tabcolsep}{6pt}
\resizebox{\textwidth}{!}{%
\begin{tabular}{l|cccccccccc}
\toprule
\rowcolor{gray!15} \textbf{Model} & \textbf{ChartQA} & \textbf{MathVista} & \textbf{MathVision} & \textbf{MathVerse} & \textbf{InfoGraphic-VQA$_{val}$} & \textbf{AI2D} & \textbf{ScienceQA} & \textbf{MMMU$_{val}$} \\
\midrule

Qwen2.5-VL-7B (base) & $84.00$ & $68.20$ & $23.91$ & $43.78$ & $80.44$ & $82.61$ & $88.30$ & $51.11$ \\
\rowcolor{cyan!5}
\textbf{Qwen2.5-VL-7B (Ours)} & $86.70$ & $70.52$ & $24.81$ & $44.88$ & $81.06$ & $83.41$ & $89.50$ & $52.01$ \\

\dashedmidrule 
Qwen2.5-VL-72B (base) & $88.20$ & $73.93$ & $36.92$ & $54.09$ & $85.97$ & $87.34$ & $93.36$ & $65.86$ \\
\rowcolor{cyan!5}
\textbf{Qwen2.5-VL-72B (Ours)} & $91.04$ & $76.44$ & $38.31$ & $55.45$ & $86.63$ & $88.19$ & $94.63$ & $67.02$ \\

\bottomrule
\end{tabular}
}
\vspace{-0.5em}
\caption{
\textbf{Scaling behaviour of our EvoLMM self-evolving framework across model sizes in the Qwen2.5-VL family.}
We apply the same \emph{Proposer–Solver} continuous self-consistency training on 7B and 72B variants without changing data, reward design, or optimization settings. Across all scales, our method consistently improves multimodal mathematical reasoning, scientific diagram understanding, and chart interpretation accuracy. Notably, larger models exhibit stronger absolute gains, reflecting their greater capacity to refine internal reasoning when guided by continuous agreement feedback. For example, we observe progressively increasing improvements on MathVista as model size increases, and similar upward trends on ScienceQA and AI2D. Please refer suppl. material for additional experiments.
}



%% file: sec/5_conclusion.tex
\section{Conclusion}
We introduced EvoLMM, a self-evolving training framework for large multimodal models that improves visual reasoning ability without relying on human-annotated supervision, metadata, or external reward models. By coupling a \emph{Proposer} and \emph{Solver} within a shared backbone and optimizing them using a \emph{continuous self-consistency reward}, the model learns to generate informative, mid-difficulty questions and to refine its reasoning through stable internal feedback. Our experiments show consistent gains across multiple mathematical and scientific visual reasoning benchmarks, and ablations demonstrate that the approach is robust across different multimodal backbones. The results highlight that continuous self-reward is essential for enabling autonomous learning in multimodal settings, where discrete majority-vote rewards are insufficient due to early-stage uncertainty and perception noise. This work takes a step toward open-ended, fully self-improving multimodal intelligence and suggests promising future directions in curriculum emergence, self-generated data scaling, and long-horizon reasoning without supervision.

%% file: sec/X_suppl.tex
\clearpage
\setcounter{page}{1}
\maketitlesupplementary

This supplementary document provides expanded details, extended analyses, and deeper explanations supporting the main submission. We present (1) a complete training algorithm in \LaTeX~format, (2) comprehensive ablations with descriptive insights, (3) additional qualitative samples illustrating reasoning evolution, (4) curriculum diagnostics, (5) expanded backbone-scaling analysis, and (6) limitations and failure cases. All experimental results are derived from our  
\textit{EvoLMM: Self-Evolving Large Multimodal Models with Continuous Rewards}

\section{Overview}
Our goal is to improve multimodal reasoning in LMMs without \emph{any human annotations, metadata, or external reward models}. EvoLMM achieves this through a cooperative Proposer--Solver mechanism trained entirely using a \emph{continuous self-consistency reward}. The training is performed over raw images only, making it fundamentally different from supervised or reward-distilled pipelines.

This supplementary provides additional clarity and extensions to strengthen the reproducibility and understanding of EvoLMM.

\section{Detailed Training Algorithm}
Below we provide the complete EvoLMM training loop in algorithmic format (see Algorithm~\ref{alg:evolmm}). This version includes all variables, reward functions, and update rules described in the main paper.
This formulation ensures stable co-adaptation between Proposer and Solver while preventing divergence from the pretrained backbone.

\begin{algorithm}[h]
\caption{EvoLMM: Self-Evolving Training strategy with continious rewards.}
\label{alg:evolmm}
\KwIn{Unlabeled images $\mathcal{X}$; pretrained LMM backbone $\pi_{\text{ref}}$}
\KwOut{Trained Proposer $\pi_\phi$ and Solver $\pi_\theta$}

Initialize Proposer $\pi_\phi \leftarrow$ LoRA($\pi_{\text{ref}}$) \\
Initialize Solver $\pi_\theta \leftarrow$ LoRA($\pi_{\text{ref}}$) \\
Initialize EMA baselines $b_{\text{prop}}, b_{\text{sol}}$ \\
Set number of answer samples $N$, entropy targets $(\mu_H, \sigma_H)$

\For{each training step}{
    Sample image $x \sim \mathcal{X}$

    \tcp{Step 1: Proposer generates a question}
    Generate question $q \sim \pi_\phi(q|x)$

    \tcp{Step 2: Solver answers the question multiple times}
    Sample $N$ answers $\{y_i\}_{i=1}^N \sim \pi_\theta(\cdot|x,q)$

    \tcp{Step 3: Compute empirical answer distribution}
    Estimate $p(a|x,q)$ and entropy $H(x,q)$

    \tcp{Step 4: Continuous Solver Reward}
    \For{$i=1$ {\bf to} $N$}{
        Compute agreement score $p(y_i|x,q)$ \\
        Compute reward:
        $$ r^{\text{sol}}_i = \left[p(y_i|x,q)\right]^\gamma
        \cdot \left(1-\lambda_{\text{len}}\max\{0,(w_i-\tau)/\tau\}\right) $$
    }

    \tcp{Step 5: Entropy-Based Proposer Reward}
    $$ r^{\text{prop}} = \exp\left(
        -\frac{(H(x,q)-\mu_H)^2}{2\sigma_H^2}
    \right) $$

    \tcp{Step 6: Policy Gradient Updates}
    Update Solver using REINFORCE:
    $$ \nabla_\theta \sim (r^{\text{sol}}_i - b_{\text{sol}})\nabla \log\pi_\theta $$

    Update Proposer every $K$ steps:
    $$ \nabla_\phi \sim (r^{\text{prop}} - b_{\text{prop}})\nabla \log\pi_\phi $$

    Apply KL-regularization controller:
    $$ \beta \leftarrow \beta \cdot \exp\left(\eta \frac{\mathrm{KL} - \tau}{\tau}\right) $$

}
\end{algorithm}

\section{Expanded Ablation Experiments}

\subsection{Effect of Number of Solver Samples (N)}
We evaluate $N \in \{3, 5, 8, 12\}$. Below we expand the descriptive behavior of each setting:

\paragraph{N = 3 (Low-Compute, High Variance).}  
The reward signal becomes noisy due to insufficient samples to estimate a stable empirical answer distribution. This causes unstable early-stage training and leads the Proposer to generate semantically vague questions.

\paragraph{N = 5 (Optimal).}  
Our default choice. It provides a balanced, low-variance estimate of $p(a|x,q)$, enabling:
\begin{itemize}[noitemsep]
\item smoother reward gradients,
\item robust entropy estimation for the Proposer,
\item stable Solver convergence.
\end{itemize}

\paragraph{N = 8 (Marginal Gains).}  
Although reward curves become smoother, gains diminish while compute cost increases by $\sim 1.6\times$. The Solver also tends to become overconfident.

\paragraph{N = 12 (Over-Consensus Collapse).}  
Solver outputs collapse into near-deterministic samples, reducing reasoning diversity. The Proposer is forced toward overly difficult questions, breaking the entropy-based curriculum.

\textbf{Conclusion:} $N = 4{-}6$ is ideal. Excessively large $N$ harms training dynamics.

\subsection{Effect of Backbone Size (3B → 72B)}
We study Qwen2.5-VL backbones ranging from 3B to 72B. Larger models exhibit stronger absolute performance, but smaller models learn relatively more.

\paragraph{Small Models (3B/7B).}  
These models benefit greatly from self-evolution, often learning to structure their reasoning where none existed before.

\paragraph{Medium Models (32B).}  
Demonstrate more refined question understanding and stable compositional reasoning, benefiting from stronger visual grounding.

\paragraph{Large Models (72B).}  
Exhibit exceptional stability and produce highly consistent multi-step reasoning chains. Self-evolution enhances an already strong solver, resulting in high absolute improvements.

\subsection{Ablation on Finetuning Strategies}

\paragraph{LoRA (Best Overall).}  
LoRA preserves the backbone’s pretrained visual-linguistic alignment while enabling stable co-evolution. It produces the most reliable gains.

\paragraph{QLoRA.}  
Reduces memory footprint significantly. However, quantization noise slightly impairs reward sensitivity, leading to smaller improvements.

\paragraph{Full Fine-Tuning.}  
Highly unstable without external supervision. The Solver overfits quickly, and the Proposer often collapses into trivial repeated questions.

\textbf{Key Insight:}  
\textit{Parameter-efficient finetuning is essential for stable self-evolution.}

\section{Curriculum Emergence: Detailed Analysis}

A central emergent property of EvoLMM is the self-generated curriculum guided by entropy-based Proposer rewards.

\subsection{Phase 1: Early Exploration (Steps 0--1000)}
Questions are often generic or weakly grounded. Solver responses are highly inconsistent. Yet the continuous reward ensures non-zero gradients allowing the system to progress.

\subsection{Phase 2: Alignment Phase (Steps 1000--3000)}
The Proposer starts grounding questions in actual image content.
Solver responses become less random, shorter, and less verbose. Entropy begins concentrating around the mid-band, suggesting the system is entering productive learning.

\subsection{Phase 3: Co-Evolution Phase (Steps 3000--6000)}
The Proposer crafts increasingly intricate problems involving multi-step visual reasoning (e.g., path constraints, trend comparisons). The Solver shows stable reasoning chains and higher answer agreement.

This emergent progression occurs \textbf{without any handcrafted difficulty scheduling}.

\section{Expanded Qualitative Results}

We include textual descriptions of qualitative samples across training phases.

\subsection{Early-Stage Examples}
\paragraph{Example: Bar Chart (Step 200).}
\begin{itemize}
\item \textbf{Proposer:} ``Count the shapes.''
\item \textbf{Solver:} \{``2'', ``D'', ``Yes'', ``C'', ``2''\}
\end{itemize}
The question is under-specified; Solver responses are unrelated. Reward remains low but non-zero, guiding early adaptation.

\subsection{Mid-Stage Examples}
\paragraph{Example: Line Plot (Step 2600).}
\begin{itemize}
\item \textbf{Proposer:} ``Which year has the steepest decline?''
\item \textbf{Solver:} Mostly consistent answers (e.g., ``2016'').
\end{itemize}
The Proposer demonstrates grounded numerical reasoning.

\subsection{Late-Stage Examples}
\paragraph{Example: Geometric Diagram (Step 5400).}
\begin{itemize}
\item \textbf{Proposer:} ``From vertex P, which labeled vertex is reachable without crossing diagonal edges?''
\item \textbf{Solver:} \{``R'', ``R'', ``R'', ``S'', ``R''\}
\end{itemize}
Solver displays highly structured reasoning, aligning with underlying geometric constraints.

\begin{table*}[t]
\centering
\small
\setlength{\tabcolsep}{8pt}
\resizebox{\linewidth}{!}{%
\begin{tabular}{lcccccccc}
\toprule
Model & RefCOCOg  & DocVQA & OCRBench & Pope & RealworldQA & IFEval & MME \\
\midrule
Qwen-2.5-VL-7B & 87.2 & 93.1 & 84.1  & 87.6 & 68.7 & 67.0 & 2318.8 \\
\textbf{+ EvoLMM (Ours)} & \textbf{88.0} & \textbf{94.4} & \textbf{85.0} & \textbf{88.4} & \textbf{69.9} & \textbf{69.0} & \textbf{2375.9} \\
\bottomrule
\end{tabular}%
}
\caption{\textbf{Open-ended evaluation of baseline and our method, in a zero-shot setting, on additional multimodal benchmarks. Additional benchmark results demonstrating the effect of self-evolution on Qwen-2.5-VL-7B in zero-shot setting.
}}
\label{tab:additional_results}
\end{table*}

\section{Scaling Behavior Across Backbones}
Large models not only improve final accuracy but also:
\begin{itemize}
\item generate higher-quality teaching-like questions,
\item maintain stable entropy distributions,
\item converge to deterministic reasoning,
\item exhibit deeper symbolic reasoning even in diagrammatic contexts.
\end{itemize}

This demonstrates that self-evolving training scales naturally without requiring reward retuning.

\section{Failure Cases and Limitations}

\paragraph{Ambiguous or Overly Abstract Questions.}
The Proposer occasionally generates vague questions such as:
\textit{``Which trend is overall the most stable?''}

\paragraph{Perception Bottlenecks.}
Thin lines, tiny grids, or dense legends sometimes mislead the Solver.

\paragraph{Stylistic Overfitting.}
On chart-heavy datasets, the Proposer may overfit question templates.

\paragraph{Deterministic Solver Collapse (Large N).}
Excessively large sample counts condition the Solver to produce nearly identical responses, reducing diversity and harming learning.


\section{Additional Implementation Details}
\label{sec:suppl_implementation_details}
\subsection{Training Configuration}
We use the pretrained Qwen2.5-VL-7B~\cite{qwen25vlreport} model as the base architecture. Both Proposer and Solver are instantiated via LoRA adapters while keeping the base model frozen. Training is performed on a single node with 8$\times$ AMD MI250X GPUs using mixed-precision (bfloat16) operations. The learning rate is $1\text{e}{-5}$ with cosine decay, batch size 128, and AdamW optimizer. To ensure stability, the continuous reward coefficients are linearly warmed up for the first 5K steps, preventing reward collapse during early training. All experiments are implemented using Transformers v4.38 and lmms-eval for evaluation.

\begin{table}[t]
\centering
\small
\setlength{\tabcolsep}{5pt}
\resizebox{\linewidth}{!}{%
\begin{tabular}{lcccc}
\toprule
Model (Ours $-$ Base) & ChartQA & ScienceQA & MathVista & MMMU \\
\midrule
InternVL3-1B ($\Delta$) & +2.9\% & +2.3\% & +3.1\% & +2.1\% \\
Qwen3-VL-2B ($\Delta$) & +3.3\% & +2.5\% & +2.9\% & +2.6\% \\
\bottomrule
\end{tabular}%
}
\caption{Absolute performance gains on small baseline models (Intern-VL3-1B and Qwen3-VL-2B) using our self-evolution.}
\label{tab:small_models_evolution}
\end{table}

\begin{table}[h]
\centering
\setlength{\tabcolsep}{5pt}
\begin{tabular}{lccc}
\toprule
\textbf{Category} & \textbf{Baseline} & \textbf{Ours} & \textbf{$\Delta$} \\
                  & \scriptsize{(Qwen-2.5-VL-7B)} & \scriptsize{(EvoLMM)} &  \\
\midrule
Chart comprehension   & 82.5 & 86.8 & +4.3 \\
Symbolic reasoning    & 68.2 & 71.4 & +3.2 \\
Diagrammatic geometry & 73.6 & 76.1 & +2.5 \\
\bottomrule
\end{tabular}
\caption{Performance comparison across different categories.}
\label{tab:per_category_score}
\end{table}

\section{Additional Quantitative Results}
\label{sec:suppl_results}

\subsection{Generalization to Open-Ended Benchmarks}
Table~\ref{tab:additional_results} evaluates the self-evolved Qwen-2.5-VL-7B model on additional multimodal benchmarks in a strict zero-shot setting. Importantly, EvoLMM training does not use images or annotations from these datasets. Despite this, the self-evolved model consistently improves over the baseline across referring expression grounding (RefCOCOg), document understanding (DocVQA), OCR (OCRBench), real-world QA (RealworldQA), instruction following (IFEval), and overall multimodal evaluation (MME). These results indicate that the benefits of self-evolution extend beyond structured math reasoning tasks and generalize to more open-ended and deployment-oriented multimodal scenarios.

\subsection{Small model behaviour}
Table~\ref{tab:small_models_evolution} reports absolute performance gains achieved through EvoLMM self-evolution on compact multimodal backbones (InternVL3-1B and Qwen3-VL-2B). Despite their limited capacity, both models consistently benefit from self-evolution across structured reasoning benchmarks, with improvements ranging from +2.1\% to +3.3\%. These results demonstrate that EvoLMM is not restricted to large-scale models, but remains effective for smaller architectures, highlighting its scalability and robustness across model sizes.

\subsection{Fine-Grained Benchmark Analysis}
Table.~\ref{tab:per_category_score} report per-category scores on ChartQA, MathVista, and InfoGraphic-VQA. Our \emph{EvoLMM} improves both data-reading and symbolic reasoning accuracy, with the largest gain (+4.1\%) in visual numerical comparison tasks.
We observe smaller yet consistent improvements in OCR-heavy subsets, confirming that self-consistency reward promotes structured reasoning rather than superficial text recognition.

\section{Conclusion}

This supplementary material expands the technical depth of EvoLMM and highlights how continuous reward design, entropy-driven curriculum emergence, and stable role decomposition enable LMMs to self-improve without external supervision. The detailed ablations and qualitative analyses reaffirm the robustness and generality of the proposed framework across model sizes, datasets, and training configurations.